%% file: acl2021.tex
\documentclass[11pt,a4paper]{article}
\usepackage[hyperref]{acl2021}
\usepackage{times}
\usepackage{latexsym}
\usepackage{amsmath}
\usepackage{amssymb}
\usepackage{bm}
\usepackage{graphicx}
\usepackage{xcolor}
\usepackage{listings}
\usepackage{setspace}
\usepackage[frozencache=true,cachedir=./mt]{minted}

\usepackage{booktabs}
\usepackage{subcaption}

\usepackage{microtype}

\aclfinalcopy 

\title{Scaling Deep Contrastive Learning Batch Size \\ under Memory Limited Setup}
\author{Luyu Gao$^1$, Yunyi Zhang$^2$, Jiawei Han$^2$, Jamie Callan$^1$ \\
  $^1$ Language Technologies Institute, Carnegie Mellon University\\
  $^2$ Department of Computer Science, University of Illinois Urbana-Champaign \\
  $^1$\texttt{\{luyug, callan\}@cs.cmu.edu} $~~~$  $^2$\texttt{\{yzhan238, hanj\}@illionis.edu} }
\date{}

\begin{document}
\maketitle
\input{abstract}

\input{intro}
\input{related}
\input{method}

\input{results}

\input{ext}
\input{conclusion}

\section*{Acknowledgments}
The authors would like to thank Zhuyun Dai and Chenyan Xiong for comments on the paper, and the anonymous reviewers for their reviews.

\clearpage
\bibliography{ref,anthology}
\bibliographystyle{acl_natbib}


\end{document}

%% file: abstract.tex
\newcommand\rurl[1]{%
  \href{http://#1}{\nolinkurl{#1}}%
}

\begin{abstract}
Contrastive learning has been applied successfully to learn vector representations of text. Previous research demonstrated that learning high-quality representations benefits from batch-wise contrastive loss with a large number of negatives. In practice, the technique of in-batch negative is used, where for each example in a batch, other batch examples' positives will be taken as its negatives, avoiding encoding extra negatives. This, however, still conditions each example's loss on all batch examples and requires fitting the entire large batch into GPU memory. This paper introduces a gradient caching technique that decouples backpropagation between contrastive loss and the encoder, removing encoder backward pass data dependency along the batch dimension. As a result, gradients can be computed for one subset of the batch at a time, leading to almost constant memory usage.
\footnote{Our code is at \rurl{github.com/luyug/GradCache}.}
\end{abstract}

%% file: intro.tex
\section{Introduction}
Contrastive learning learns to encode data into an embedding space such that related data points have closer representations and unrelated ones have further apart ones. Recent works in NLP adopt deep neural nets as encoders and use unsupervised contrastive learning on sentence representation~\cite{Giorgi2020DeCLUTRDC}, text retrieval~\cite{lee-etal-2019-latent}, and language model pre-training tasks~\cite{Wu2020CLEARCL}. Supervised contrastive learning~\cite{khosla2020supervised} has also been shown effective in training dense retrievers~\cite{karpukhin-etal-2020-dense,ding2020rocketqa}.  These works typically use batch-wise contrastive loss, sharing target texts as in-batch negatives. With such a technique, previous works have empirically shown that larger batches help learn better representations.
However, computing loss and updating model parameters with respect to a big batch require encoding all batch data and storing all activation, so batch size is limited by total available GPU memory. This limits application and research of contrastive learning methods under memory limited setup, e.g. academia. For example, \citet{lee-etal-2019-latent} pre-train a BERT~\cite{devlin-etal-2019-bert} passage encoder with a batch size of 4096 while a high-end commercial GPU RTX 2080ti can only fit a batch of 8. The gradient accumulation technique, splitting a large batch into chunks and summing gradients across several backwards, cannot emulate a large batch as each smaller chunk has fewer in-batch negatives.

In this paper, we present a simple technique that thresholds peak memory usage for contrastive learning to almost constant regardless of the batch size. For deep contrastive learning, the memory bottlenecks are at the deep neural network based encoder.
We observe that we can separate the back-propagation process of contrastive loss into two parts, from loss to representation, and from representation to model parameter, with the latter being independent across batch examples given the former, detailed in \autoref{sec:analysis}. We then show in \autoref{sec:technique} that by separately pre-computing the representations' gradient and store them in a cache, we can break the update of the encoder into multiple sub-updates that can fit into the GPU memory. 
This pre-computation of gradients allows our method to produce the \emph{exact same} gradient update as training with large batch. Experiments show that with about 20\% increase in runtime, our technique enables a single consumer-grade GPU to reproduce the state-of-the-art large batch trained models that used to require multiple professional GPUs. 

%% file: related.tex
\section{Related Work}

\paragraph{Contrastive Learning}
First introduced for probablistic language modeling~\cite{Mnih2012AFA}, Noise Contrastive Estimation~(NCE) was later used by Word2Vec~\cite{Mikolov2013DistributedRO} to learn word embedding. Recent works use contrastive learning to unsupervisedly pre-train~\cite{lee-etal-2019-latent,ChangYCYK20} as well as supervisedly train dense retriever~\cite{karpukhin-etal-2020-dense}, where contrastive loss is used to estimate retrieval probability over the entire corpus. Inspired by SimCLR~\cite{Chen2020ASF}, constrastive learning is used to learn better sentence representation~\cite{Giorgi2020DeCLUTRDC} and pre-trained language model~\cite{Wu2020CLEARCL}.

 
 \paragraph{Deep Network Memory Reduction}
Many existing techniques deal with large and deep models. The gradient checkpoint method attempts to emulate training deep networks by training shallower layers and connecting them with gradient checkpoints and re-computation~\cite{Chen2016TrainingDN}. Some methods also use reversible activation functions, allowing internal activation in the network to be recovered throughout back propagation~\cite{Gomez2017TheRR,MacKay2018ReversibleRN}. However, their effectiveness as part of contrastive encoders has not been confirmed. 
Recent work also attempts to remove the redundancy in optimizer tracked parameters on each GPU~\cite{rajbhandari2020zero}. Compared with the aforementioned methods, our method is designed for scaling over the batch size dimension for contrastive learning. 

%% file: method.tex
\section{Methodologies}
In this section, we formally introduce the notations for contrastive loss and analyze the difficulties of using it on limited hardware. We then show how we can use a Gradient Cache technique to factor the loss so that large batch gradient update can be broken into several sub-updates.


\subsection{Preliminaries}
Under a general formulation, given two classes of data $\mathcal{S}, \mathcal{T}$, we want to learn encoders $f$ and $g$ for each such that, given $s\in\mathcal{S}, t \in \mathcal{T}$, encoded representations $f(s)$ and $g(t)$ are close if related and far apart if not related by some distance measurement.
For large $\mathcal{S}$ and $\mathcal{T}$ and deep neural network based $f$ and $g$, direct training is not tractable, so a common approach is to use a contrastive loss: sample anchors $S \subset \mathcal{S}$ and targets $T \subset \mathcal{T}$ as a training batch, where each element $s_i \in S$ has a related element $t_{r_i} \in T$ as well as zero or more specially sampled hard negatives. 
The rest of the random samples in $T$ will be used as in-batch negatives. Define loss based on dot product as follows:
\begin{equation}
    \mathcal{L} 
    = - \frac{1}{|S|} \sum_{s_i \in S} \log 
    \frac{exp(f(s_i)^\top g(t_{r_i}) / \tau)}
    {\sum_{t_j \in T} exp(f(s_i)^\intercal g(t_j) / \tau)}
\end{equation}
where each summation term depends on the \emph{entire} set $T$ and requires fitting \emph{all} of them into memory. 

We set temperature
$\tau = 1$ in the following discussion for simplicity as in general it only adds a constant multiplier to the gradient. 

\subsection{Analysis of Computation}
\label{sec:analysis}
In this section, we give a mathematical analysis of contrastive loss computation and its gradient. We show that the back propagation process can be divided into two parts, from loss to representation, and from representation to encoder model.  The separation then enables us to devise a technique that removes data dependency in encoder parameter update. 
Suppose the function $f$ is parameterized with $\Theta$ and $g$ is parameterized with $\Lambda$.
\begin{align}
    \frac{\partial \mathcal{L}}{\partial \Theta} &= \sum_{s_i \in S} \frac{\partial \mathcal{L}}{\partial f(s_i)} \frac{\partial f(s_i)}{\partial \Theta} \\
    \frac{\partial \mathcal{L}}{\partial \Lambda} &= \sum_{t_j \in T} \frac{\partial \mathcal{L}}{\partial g(t_j)} \frac{\partial g(t_j)}{\partial \Lambda}
\end{align}
As an extra notation, denote normalized similarity,
\begin{equation}
   p_{ij} = \frac{exp(f(s_i)^\intercal g(t_j))}{\sum_{t \in T} exp(f(s_i)^\intercal g(t))}
\end{equation}
We note that the summation term for a particular $s_i$ or $t_i$ is a function of the batch, as,
\begin{align}
    \frac{\partial \mathcal{L}}{\partial f(s_i)} &= - \frac{1}{|S|}\left( g(t_{r_i}) - \sum_{t_j \in T} p_{ij} g(t_j) \right),\\
    \frac{\partial \mathcal{L}}{\partial g(t_j)} &= - \frac{1}{|S|}\left( \epsilon_j - \sum_{s_i \in S} p_{ij} f(s_i) \right),
    \label{eq:rep-grad}
\end{align}
where
\begin{equation}
    \epsilon_j =
    \begin{cases}
    f(s_k) & \text{if} \; \exists\; k \; \text{s.t.} \; r_k = j \\
    0   & \text{otherwise}
\end{cases}
\end{equation}
which prohibits the use of gradient accumulation.
We make two observations here:
\begin{itemize}
    \item 
    The partial derivative $\frac{\partial f(s_i)}{\partial \Theta}$ depends only on $s_i$ and $\Theta$ while $\frac{\partial g(t_j)}{\partial \Lambda}$ depends only on $t_j$ and $\Lambda$; and
    \item 
    Computing partial derivatives $\frac{\partial \mathcal{L}}{\partial f(s_i)}$ and $\frac{\partial \mathcal{L}}{\partial g(t_j)}$ requires only encoded representations, but not $\Theta$ or $\Lambda$.
\end{itemize}
These observations mean back propagation of $f(s_i)$ for data $s_i$ can be run independently with its own computation graph and activation if the \emph{numerical} value of the partial derivative $\frac{\partial \mathcal{L}}{\partial s_i}$ is known. Meanwhile the derivation of $\frac{\partial \mathcal{L}}{\partial s_i}$ requires only \emph{numerical} values of two sets of representation vectors $F = \{f(s_1), f(s_2), .., f(s_{|S|})\}$ and $G=\{g(t_1), g(t_2), ..., g(t_{|T|})\}$. A similar argument holds true for $g$, where we can use representation vectors to compute $\frac{\partial \mathcal{L}}{\partial t_j}$ and back propagate for each $g(t_j)$ independently. In the next section, we will describe how to scale up batch size by pre-computing these representation vectors.

\subsection{Gradient Cache Technique}
\label{sec:technique}
Given a large batch that does not fit into the available GPU memory for training, we first divide it into a set of sub-batches each of which can fit into memory for gradient computation, denoted as $\mathbb{S} = \{\hat{S}_1, \hat{S}_2, ..\}, \mathbb{T}=\{\hat{T}_1, \hat{T}_2, ..\}$. The full-batch gradient update is computed by the following steps. 
\paragraph{Step1: Graph-less Forward}
Before gradient computation, we first run an extra encoder forward pass for each batch instance to get its representation. Importantly, this forward pass runs without constructing the computation graph. We collect and store all representations computed.
\paragraph{Step2: Representation Gradient Computation and Caching}
We then compute the contrastive loss for the batch based on the representation from Step1 and have a corresponding computation graph constructed. Despite the mathematical derivation, automatic differentiation system is used in actual implementation, which automatically supports variations of contrastive loss.
A backward pass is then run to populate gradients for each representation. Note that the encoder is not included in this gradient computation. Let $\mathbf{u}_i = \frac{\partial \mathcal{L}}{\partial f(s_i)}$ and $\mathbf{v}_i = \frac{\partial \mathcal{L}}{\partial g(t_i)}$, we take these gradient tensors and store them as a~\emph{Representation Gradient Cache}, $[\mathbf{u}_1, \mathbf{u}_2, .., \mathbf{v}_1, \mathbf{v}_2, ..]$.

\paragraph{Step3: Sub-batch Gradient Accumulation}
We run encoder forward one sub-batch at a time to compute representations and build the corresponding computation graph. We take the sub-batch's representation gradients from the cache and run back propagation through the encoder. Gradients are accumulated for encoder parameters across all sub-batches. Effectively for $f$ we have,
\begin{align}
\begin{split}
    \frac{\partial \mathcal{L}}{\partial \Theta} 
    &=  \sum_{\hat{S}_j \in \mathbb{S}} \sum_{s_i \in \hat{S}_j} \frac{\partial \mathcal{L}}{\partial f(s_i)} \frac{\partial f(s_i)}{\partial \Theta} \\
    &=  \sum_{\hat{S}_j \in \mathbb{S}} \sum_{s_i \in \hat{S}_j} \mathbf{u}_i \frac{\partial f(s_i)}{\partial \Theta}
\end{split}
\end{align}
where the outer summation enumerates each sub-batch and the entire internal summation corresponds to one step of accumulation.
Similarly, for $g$, gradients accumulate based on,
\begin{equation}
    \frac{\partial \mathcal{L}}{\partial \Lambda} 
    = \sum_{\hat{T}_j \in \mathbb{T}} \sum_{t_i \in \hat{T}_j} \mathbf{v}_i \frac{\partial g(t_i)}{\partial \Lambda}
\end{equation}
Here we can see the \emph{equivalence} with direct large batch update by combining the two summations.
\paragraph{Step4: Optimization} When all sub-batches are processed, we can step the optimizer to update model parameters as if the full batch is processed in a single forward-backward pass.

Compared to directly updating with the full batch, which requires memory linear to the number of examples, our method fixes the number of examples in each encoder gradient computation to be the size of sub-batch and therefore requires constant memory for encoder forward-backward pass. The extra data pieces introduced by our method that remain persistent across steps are the representations and their corresponding gradients with the former turned into the latter after representation gradient computation. Consequently, in a general case with data from $S$ and $T$ each represented with $d$ dimension vectors, we only need to store $(|S|d + |T|d)$ floating points in the cache on top of the computation graph. To remind our readers, this is several orders smaller than million-size model parameters.

\subsection{Multi-GPU Training}
\label{sec:multi}
When training on multiple GPUs, we need to compute the gradients with all examples across all GPUs. This requires a single additional cross GPU communication after \emph{Step1} when all representations are computed. We use an all-gather operation to make all representations available on all GPUs. Denote $F^n, G^n$ representations on $n$-th GPU and a total of N device.  \emph{Step2} runs with gathered representations $F^\text{all} = F^1 \cup .. \cup F^N$ and $G^\text{all} = G^1 \cup .. \cup G^N$. While $F^\text{all}$ and $G^\text{all}$ are used to compute loss, the $n$-th GPU only computes gradient of its local representations $F^n, G^n$ and stores them into cache. No communication happens in \emph{Step3}, when each GPU independently computes gradient for local representations. \emph{Step4} will then perform gradient reduction across GPUs as with standard parallel training.

%% file: results.tex
\section{Experiments}
To examine the reliability and computation cost of our method, we implement our method into dense passage retriever~(DPR; \citet{karpukhin-etal-2020-dense})\footnote{Our implementation is at: \url{https://github.com/luyug/GC-DPR}}. We use gradient cache to compute DPR's supervised contrastive loss on a single GPU. Following DPR paper, we measure top hit accuracy on the Natural Question Dataset~\cite{Kwiatkowski2019NaturalQA} for different methods. We then examine the training speed of various batch sizes.

\begin{table}
\centering
\begin{tabular}{l | c c c}
\toprule
 Method & Top-5 & Top-20 & Top-100    \\
 \midrule
 DPR & - & 78.4 & 85.4 \\
 \midrule
 Sequential & 59.3 & 71.9 & 80.9 \\
 Accumulation & 64.3 & 77.2 & 84.9\\
 \midrule
 Cache & 68.6 & 79.3 & 86.0 \\
 - BSZ = 512 & 68.3 & 79.9 & 86.6 \\
 \bottomrule
\end{tabular}
\vspace{-2mm}
\caption{Retrieval: We compare top-5/20/100 hit accuracy of small batch update~(Sequential), accumulated small batch~(Accumulation) and gradient cache~(Cache) systems with DPR reference.}
\label{tab:acc}
\vspace{-3mm}
\end{table}

\subsection{Retrieval Accuracy}
\paragraph{Compared Systems} 1) \textbf{DPR}: the reference number taken from the original paper trained on 8 GPUs, 2) \textbf{Sequential}: update with max batch size that fits into 1 GPU, 3) \textbf{Accumulation}: similar to Sequential but accumulate gradients and update until number of examples matches DPR setup, 4)~\textbf{Cache}: training with DPR setup using our gradient cache on 1 GPU. 
We attempted to run with gradient checkpointing but found it cannot scale to standard DPR batch size on our hardware.

\paragraph{Implementations} All runs start with the same random seed and follow DPR training hyperparameters except batch size. Cache uses a batch size of 128 same as DPR and runs with a sub-batch size of 16 for questions and 8 for passages. We also run Cache with a batch size of 512~(BSZ=512) to examine the behavior of even larger batches. Sequential uses a batch size of 8, the largest that fits into memory. Accumulation will accumulate 16 of size-8 batches. Each question is paired with a positive and a BM25 negative passage. All experiments use a single RTX 2080ti.

\paragraph{Results} Accuracy results are shown in ~\autoref{tab:acc}. We observe that Cache performs better than DPR reference due to randomness in training. Further increasing batch size to 512 can bring in some advantage at top 20/100. Accumulation and Sequential results confirm the importance of a bigger batch and more negatives. For Accumulation which tries to match the batch size but has fewer negatives, we see a drop in performance which is larger towards the top. In the sequential case, a smaller batch incurs higher variance, and the performance further drops. In summary, our Cache method improves over standard methods and matches the performance of large batch training.

\begin{figure}[t]
  \centering
  \includegraphics[width=0.47\textwidth]{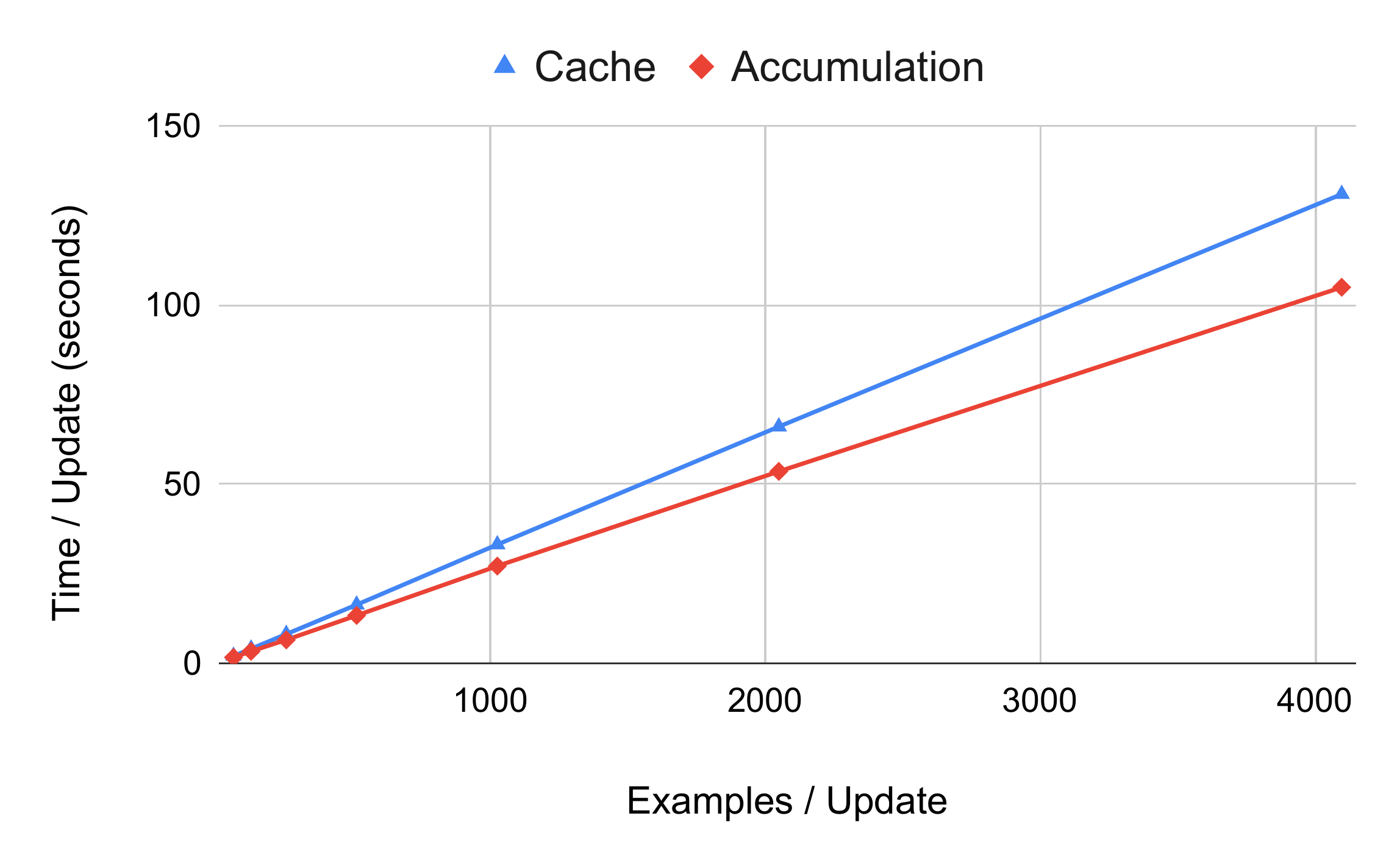}
  \vspace{-5mm}
  \caption{We compare training speed versus the number of examples per update for gradient cache (Cache) and gradient accumulation (Accumulation).}
  \label{fig:scale}
  \vspace{-0.5cm}
\end{figure}

\subsection{Training Speed}
In \autoref{fig:scale}, we compare update speed of gradient cache and accumulation with per update example number of $\{64, 128, 256, 512, 1024, 2048, 4096\}$. We observe gradient cache method can steadily scale up to larger batch update and uses 20\% more time for representation pre-computation. This extra cost enables it to create an update of a much larger batch critical for the best performance, as shown by previous experiments and many early works. While the original DPR reports a training time of roughly one day on 8 V100 GPUs, in practice, with improved data loading, our gradient cache code can train a dense retriever in a practical 31 hours on a single RTX2080ti. We also find gradient checkpoint only runs up to batch of 64 and consumes twice the amount of time than accumulation\footnote{We used the gradient checkpoint implemented in Huggingface transformers package}.

%% file: conclusion.tex
\section{Extend to Deep Distance Function}

Previous discussion assumes a simple parameter-less dot product similarity. In general it can also be deep distance function $\Phi$ richly parameterized by $\Omega$, formally,
\begin{equation}
    d_{ij} = d(s_i, t_j) = \Phi (f(s_i), g(t_j))
\end{equation}
This can still scale by introducing an extra \emph{Distance Gradient Cache}. In the first forward we collect all representations as well as all distances. We compute loss with $d_{ij}$s and back propagate to get $w_{ij} = \frac{\partial \mathcal{L}}{\partial d_{ij}}$, and store them in Distance Gradient Cache, $[w_{00}, w_{01}, .., w_{10}, ..]$. We can then update $\Omega$ in a sub-batch manner,
\begin{equation}
    \frac{\partial \mathcal{L}}{\partial \Omega}
    = \sum_{\hat{S} \in \mathbb{S}} \sum_{\hat{T} \in \mathbb{T}} \sum_{s_i \in \hat{S}} \sum_{t_j \in \hat{T}}
    w_{ij} \frac{\partial \Phi (f(s_i), g(t_j))}{\partial \Omega}
\end{equation}
 Additionally, we \emph{simultaneously} compute with the constructed computation graph $\frac{\partial d_{ij}}{\partial f(s_i)}$ and $\frac{\partial d_{ij}}{\partial g(t_j)}$ and accumulate across batches,
\begin{equation}
    \mathbf{u}_i = \frac{\partial \mathcal{L}}{\partial f(s_i)} = \sum_j w_{ij} \frac{\partial d_{ij}}{\partial f(s_i)}
\end{equation}
and,
\begin{equation}
    \mathbf{v}_j = \frac{\partial \mathcal{L}}{\partial g(t_j)} = \sum_i w_{ij} \frac{\partial d_{ij}}{\partial g(t_j)}
\end{equation}
with which we can build up the Representation Gradient Cache. When all representations' gradients are computed and stored, encoder gradient can be computed with \emph{Step3} described in \autoref{sec:technique}. In philosophy this method links up two caches. Note this covers early interaction $f(s)=s, g(t)=t$ as a special case.

\section{Conclusion}
In this paper, we introduce a gradient cache technique that breaks GPU memory limitations for large batch contrastive learning. We propose to construct a representation gradient cache that removes in-batch data dependency in encoder optimization. Our method produces the exact same gradient update as training with a large batch. 
We show the method is efficient and capable of preserving accuracy on resource-limited hardware. 
We believe a critical contribution of our work is providing a large population in the NLP community with access to batch-wise contrastive learning. While many previous works come from people with industry-grade hardware,
researchers with limited hardware can now use our technique to reproduce state-of-the-art models and further advance the research without being constrained by available GPU memory.